\documentclass[10pt,twocolumn,letterpaper]{article}

\usepackage{wacv}
\usepackage{times}
\usepackage{epsfig}
\usepackage{graphicx}
\usepackage{amsmath}
\usepackage{amssymb}
\usepackage{booktabs}

\def\wacvPaperID{12} 

\wacvapplicationstrack 

\wacvfinalcopy 

\ifwacvfinal
\usepackage[breaklinks=true,bookmarks=false]{hyperref}
\else
\usepackage[pagebackref=true,breaklinks=true,colorlinks,bookmarks=false]{hyperref}
\fi

\begin{document}

\title{Pose Estimation for Human Wearing Loose-Fitting Clothes --Obtaining Ground Truth Posture Using HFR Camera and Blinking LEDs}

\author{Takayoshi Yamaguchi, Dan Mikami\\
Kogakuin University\\
1-24-2 Nishi-shinjuku, Shinjuku, Tokyo, Japan\\
{\tt\small em22025@ns.kogakuin.ac.jp, mikami.dan@cc.kogakuin.ac.jp}
\and
Seiji Matsumura, Naoki Saijo, Makio Kashino\\
NTT Communication Science Labratories\\
3-1 Morinosato-wakamiya, Atsugi, Kanagawa, Japan\\
{\tt\small \{seiji.matsumura.yh, naoki.saijo, makio.kashino.ft\}@hco.ntt.co.jp}
}

\maketitle
\thispagestyle{empty}

\begin{abstract}
Human pose estimation, particularly in athletes, can help improve their performance. However, this estimation is difficult using existing methods, such as human annotation, if the subjects wear loose-fitting clothes such as ski/snowboard wears. This study developed a method for obtaining the ground truth data on two-dimensional (2D) poses of a human wearing loose-fitting clothes. This method uses fast-flushing light-emitting diodes (LEDs). The subjects were required to wear loose-fitting clothes and place the LED on the target joints. The LEDs were observed directly using a camera by selecting thin filmy loose-fitting clothes. The proposed method captures the scene at 240 fps by using a high-frame-rate camera and renders two 30 fps image sequences by extracting LED-on and -off frames. The temporal differences between the two video sequences can be ignored, considering the speed of human motion. The LED-on video was used to manually annotate the joints and thus obtain the ground truth data. Additionally, the LED-off video, equivalent to a standard video at 30 fps, confirmed the accuracy of existing machine learning-based methods and manual annotations. Experiments demonstrated that the proposed method can obtain ground truth data for standard RGB videos. Further, it was revealed that neither manual annotation nor the state-of-the-art pose estimator obtains the correct position of target joints.
\end{abstract}

\section{Introduction}


The application of data analysis is becoming increasingly popular for improving sports performance \cite{paul,DeFroda,LAUDNER2019330,9592793}. Objective data, such as pose data, are expected to be a critical part of these analyses that result in innovations. In the past decade, human pose estimation has become one of the most active research fields in computer vision. The solutions vary from camera-based \cite{8765346,Moon_2018_CVPR_V2V-PoseNet,mmpose2020} and inertial-sensor-based methods \cite{6402423,marcard} to hybrid of these methods. Inertial-sensor-based methods require the attachment of small sensors to the body. Subsequently, postural data can be obtained by analyzing the sensors' output. This wearable sensor-based solution can be applied to a target that moves over a wide area. As wearable sensor-based methods are attached to the body, these sensors may hinder the subject, particularly in sports involving acrobatic movements or body contact. Another disadvantage is its influence on the subject's performance. Wearing a sensor restricts the athletes' routine, impacting their performance.


Conversely, camera-based methods can estimate poses without the subject's assistance. Recently, deep learning-based methods have realized automatic estimation of human poses only from camera images.



Learning-based methods, such as deep learning-based approaches, are based on training data. Thus, they cannot estimate untrained aspects. Current human pose estimators are trained using public data sources, such as MS COCO \cite{coco} and MPII \cite{andriluka14cvpr}. Some studies have focused on estimating poses without having similar situations included in the training data \cite{Kitamura_2022_WACV,Zwolfer2021Improved2K}, such as upside-down poses in pole vaulting. They used additional training data suited to their situation and improved the pose-estimation performance.



This study focused on pose estimation of subjects wearing loose-fitting clothes as a situation not included in existing datasets. Some types of winter sports, such as freestyle skiing or snowboarding, necessitate the wearing of loose-fitting clothes. Therefore, we considered the need for human pose estimation under loose-fitting clothes, which is significant for this community.



Considering previous approaches, this problem may appear to be easily resolved by including situations with loose-fitting clothes in the training data. However, obtaining the joint positions with loose-fitting clothes requires considerable effort. Even for a human annotator, determining the joint position is difficult because of the shaking of the clothes.



Matsumoto et al. estimated the pose of a human target wearing Japanese kimonos as an example of loose-fitting clothing \cite{TakuyaMATSUMOTO20202019MVP0007}. The method requires the subjects to perform the same movement twice, wearing tight- and loose-fitting clothes. They subsequently estimated the joint position occluded in loose-fitting clothes by matching these two motion sequences. According to the study, motion similarity between the two movements is essential, and accuracy deteriorates when the motion similarity decreases. Thus, this approach is unsuitable for sports applications.



The primary goal of this study was to enable 2D human pose estimation of a subject wearing loose-fitting clothes in sports scenarios. To this end, this study developed a simple but novel method to obtain the ground truth value of joints in loose-fitting clothes. The proposed method uses fast-flashing light-emitting diode (LED) lights placed on the subject's joints. The participants wore loose-fitting clothes over the lights. Filmy clothes made of linen or polyester, which adequately allow light to pass through, are selected as the loose-fitting clothes. The video was captured at 240 fps using a high-frame-rate camera. By extracting the light-on and -off frames, we rendered two video sequences of 30 fps each––one obtained the ground truth joint position from the lighted LED, and the other was used as regular video sequences.



In the experiment, we obtained data from multiple sequences of persons and motions. We obtained the ground truth data by manually clicking on a light-on video sequence rendered using the proposed method. Subsequently, a state-of-the-art (SOTA) human pose estimation method, MMPose \cite{mmpose2020}, was applied to light-off video sequences. Additionally, two operators manually labeled the joint positions in the light-off video sequences. The estimates and ground truths were compared. We verified that the pose estimates from MMPose and human operators differed from the ground truth data. Therefore, our data can help improve the human pose estimator. Moreover, the ground truth data feedback can improve the human operators' accuracy.


\section{Proposed method}


This study entailed the development of a method for obtaining human joint position ground truth data in 2D with the subject wearing loose-fitting clothes. In the proposed method, fast-flushing LED are attached onto the joints, and the video is captured using a high-frame-rate camera; it virtually enables the temporal division of the information source. The subsequent sections present the details of video capturing and post-processing.


\subsection{Capturing method}


The proposed method uses fast-flushing LEDs. The detailed setting had light-on and -off times of 10 and 23.33 ms, respectively. The device was easily implemented using the Arduino Nano and a relay. The participants, two males in their 30s and 20s, wore loose-fitting clothes. Thin polyester clothes were selected.



\begin{figure}
\begin{center}
\includegraphics[pagebox=cropbox,width=0.48\textwidth]{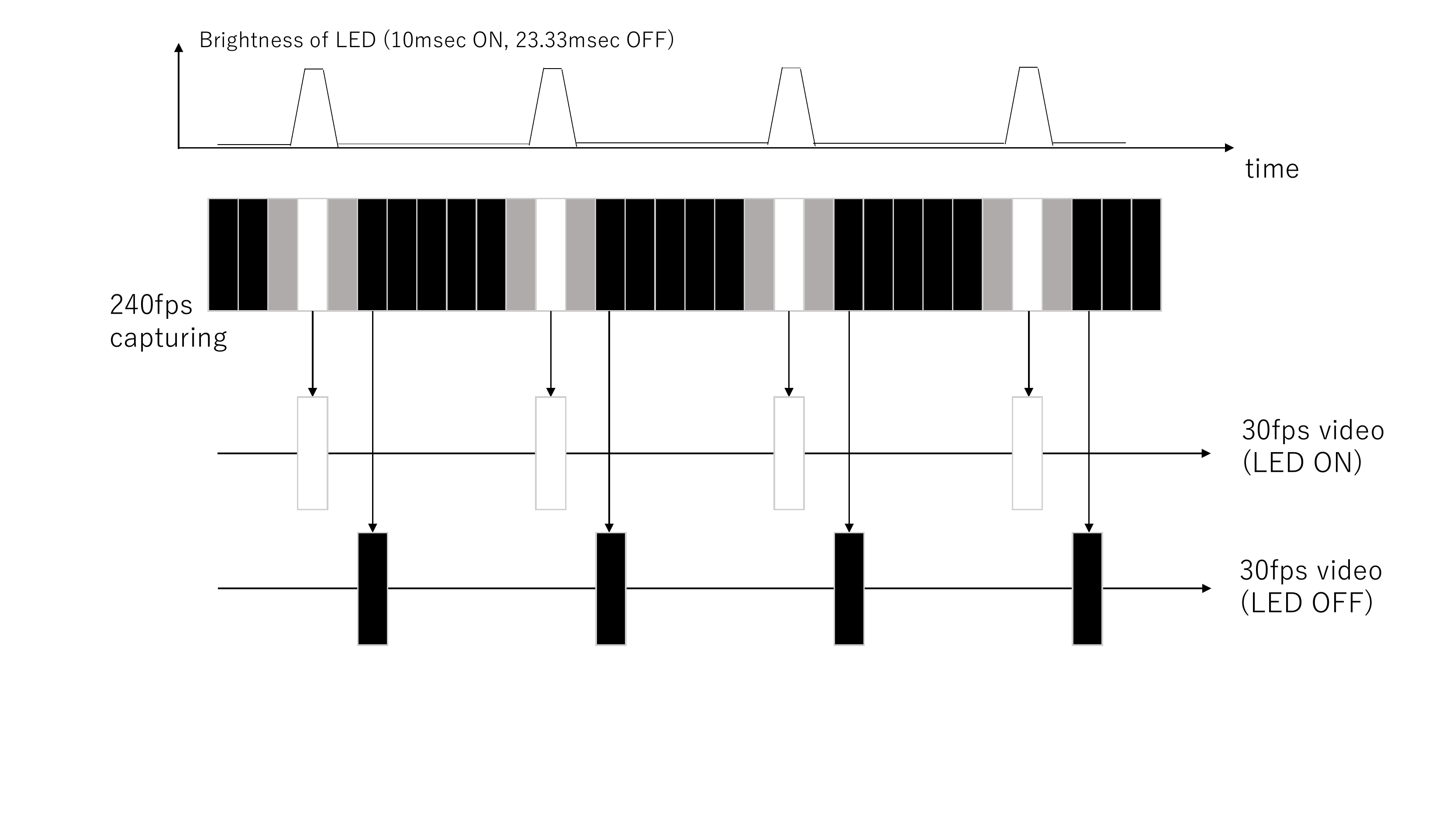}
\end{center}
\vspace{-1cm}
   \caption{Concept of the proposed approach. The topmost figure shows the brightness of the LED. It is captured at 240 fps. The middle part of the figure shows the brightness in a 240 fps video sequence. From the HFR video, LED-on and -off video sequence is extracted.}
\label{fig:fig1}
\end{figure}

\begin{figure}
\begin{center}
\includegraphics[pagebox=cropbox,width=0.26\textwidth]{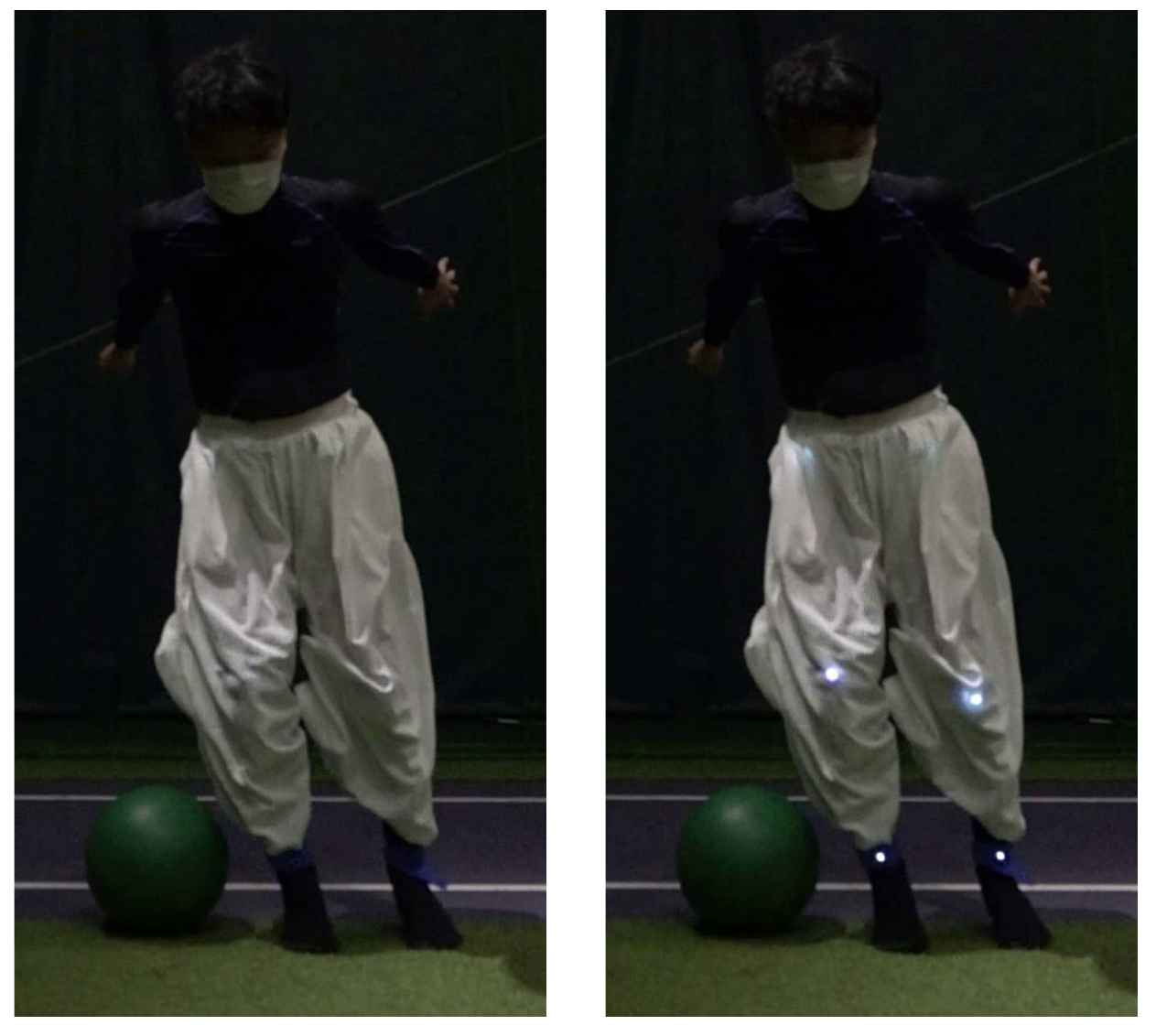}
\end{center}
   \caption{Example of the LED-off and -on image. The captured timing of the two images differs by 8.33 msec. 
 in time. However, the two images are almost the same to the extent of capturing human motion. Thus, the temporal difference can ignore.}
\label{fig:fig2}
\end{figure}

The video sequences were captured at 240 fps by using a high-frame-rate camera. The light-on and -off cycle was 33.33 ms, corresponding to eight frames on 240 fps video, as shown in Fig. \ref{fig:fig1}. Figure \ref{fig:fig2} shows sample snapshots, with Fig. \ref{fig:fig2}(a) illustrating the LED-on image and Fig. \ref{fig:fig2}(b) showing the LED-off image. As shown in Fig. \ref{fig:fig2}, considering human motion speed, the temporal difference of 2 [frames]/240 [fps] = 8.3 [ms] between (a) and (b) can be ignored.


\subsection{Post-processing}


Following the capturing of the 240-fps video, two 30 fps sequences were rendered: an LED-on video sequence and an LED-off video sequence, as shown in Fig. \ref{fig:fig1}. Each cycle of LED on and off consisted of eight frames. One LED-on frame was extracted from eight successive frames; one LED-off frame was also selected similarly. In this study, we manually extracted LED-on and -off frames. The LED position was then manually annotated by clicking on the LED-on video. Hereafter, we refer to the manually annotated data as the ground truth data.


\section{Experiment}

\begin{figure}
\begin{center}
\includegraphics[pagebox=cropbox,width=0.3\textwidth]{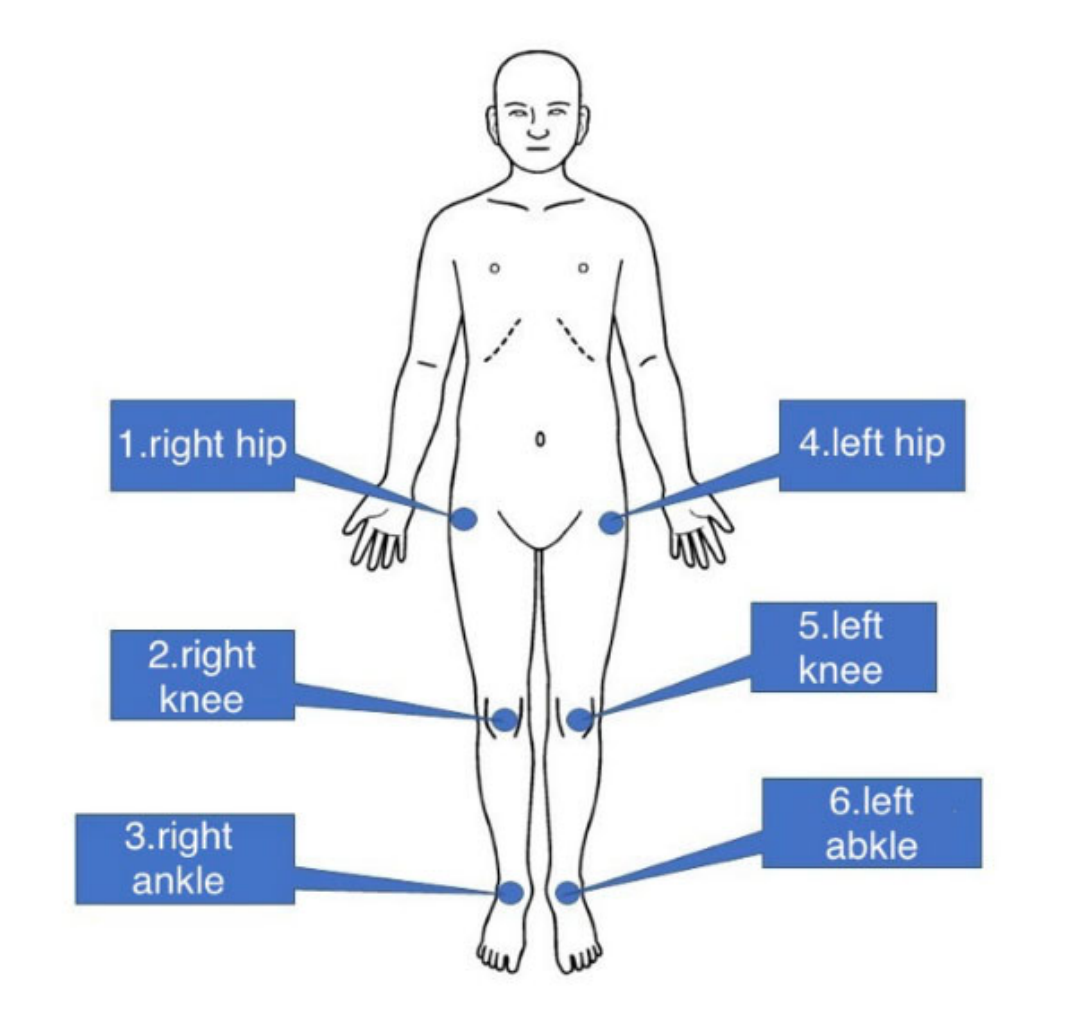}
\end{center}
   \caption{Position of LEDs. We used the six LEDs in the experiment.}
\label{fig:fig3}
\end{figure}

The effectiveness of the proposed method was experimentally verified. Two types of target movements were captured––a fundamental movement and a skiing motion on a ski simulator––from two subjects. For the experimental verification, we attached LEDs on both sides of the hip, knee, and ankle joints, as shown in Fig. \ref{fig:fig3}.


The following two comparisons were conducted. First, we compared the estimates of the joint positions of the two human operators with the ground truth. Human operators were asked to annotate the positions of the six joints on the LED-off video sequence. Second, we compared the estimates from the learning-based pose estimator with the ground truth. MMpose was used as the learning-based pose estimator and was applied to the LED-off video sequence similarly to the human operator case. 



\subsection{Motion sequences}

\begin{figure}
\begin{center}
\includegraphics[pagebox=cropbox,width=0.48\textwidth]{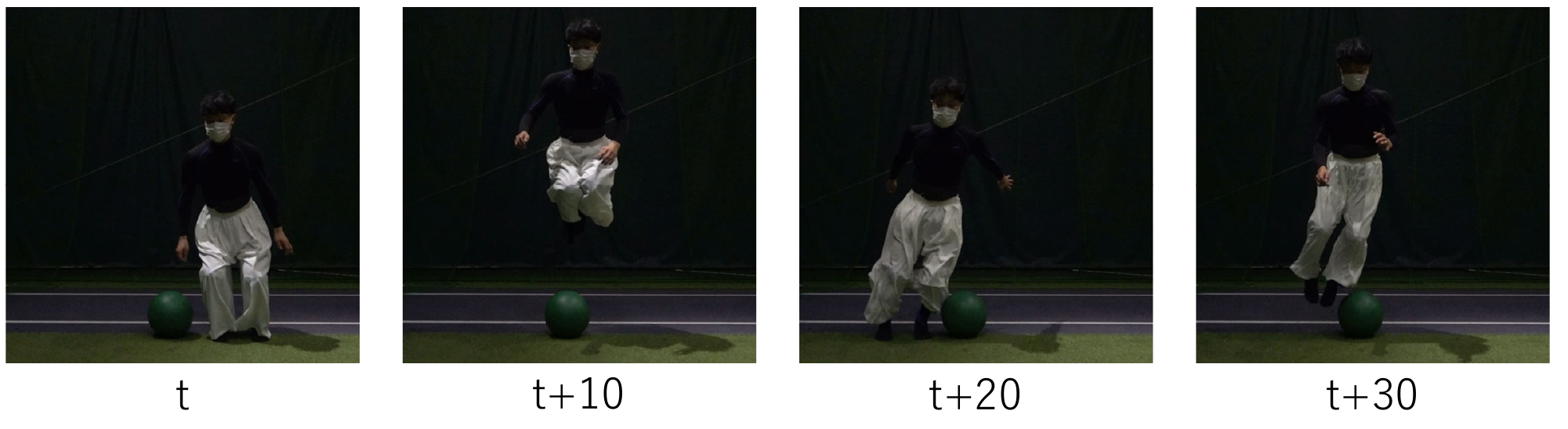}
\end{center}
   \caption{Snapshots of the sideways jump. Each image is ten frames apart in time. It continues about 160 frames.}
\label{fig:fig4}
\begin{center}
\includegraphics[pagebox=cropbox,width=0.48\textwidth]{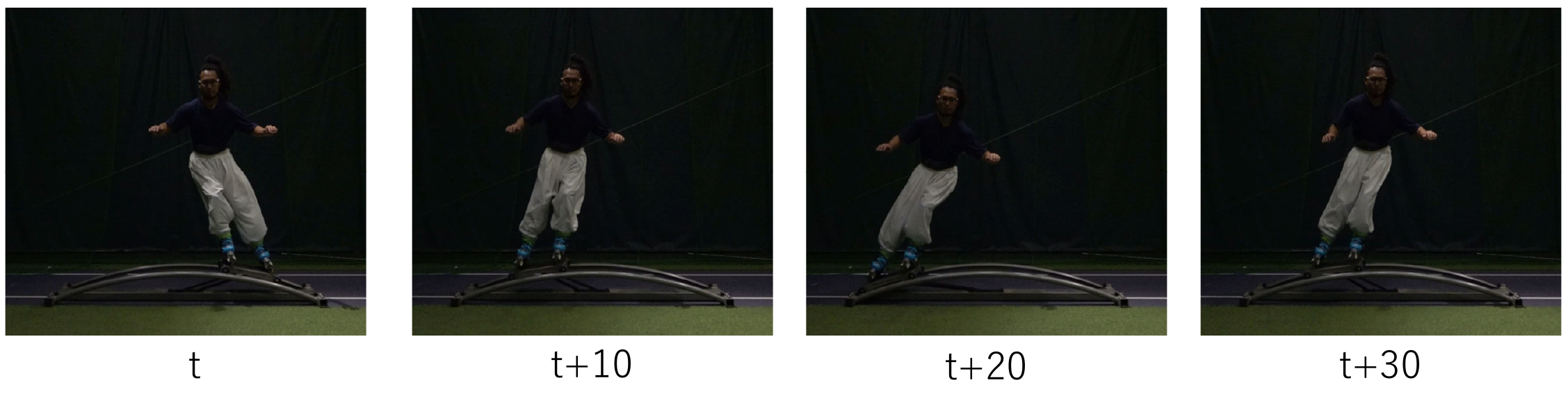}
\end{center}
   \caption{Snapshots of ski simulator motion.}
\label{fig:fig5}
\end{figure}

We used a sideways jump over a ball for the fundamental movement, as shown in Fig.\ref{fig:fig4}. Additionally, we captured the action in the ski simulator, as shown in Fig.\ref{fig:fig5}. Both movements had vertical and horizontal translations. Two subjects were asked to perform the two types of the aforementioned movements.


\subsection{Comparisons between existing methods and ground truths}

\begin{figure}
  \begin{center}
    \includegraphics[pagebox=cropbox,width=0.48\textwidth]{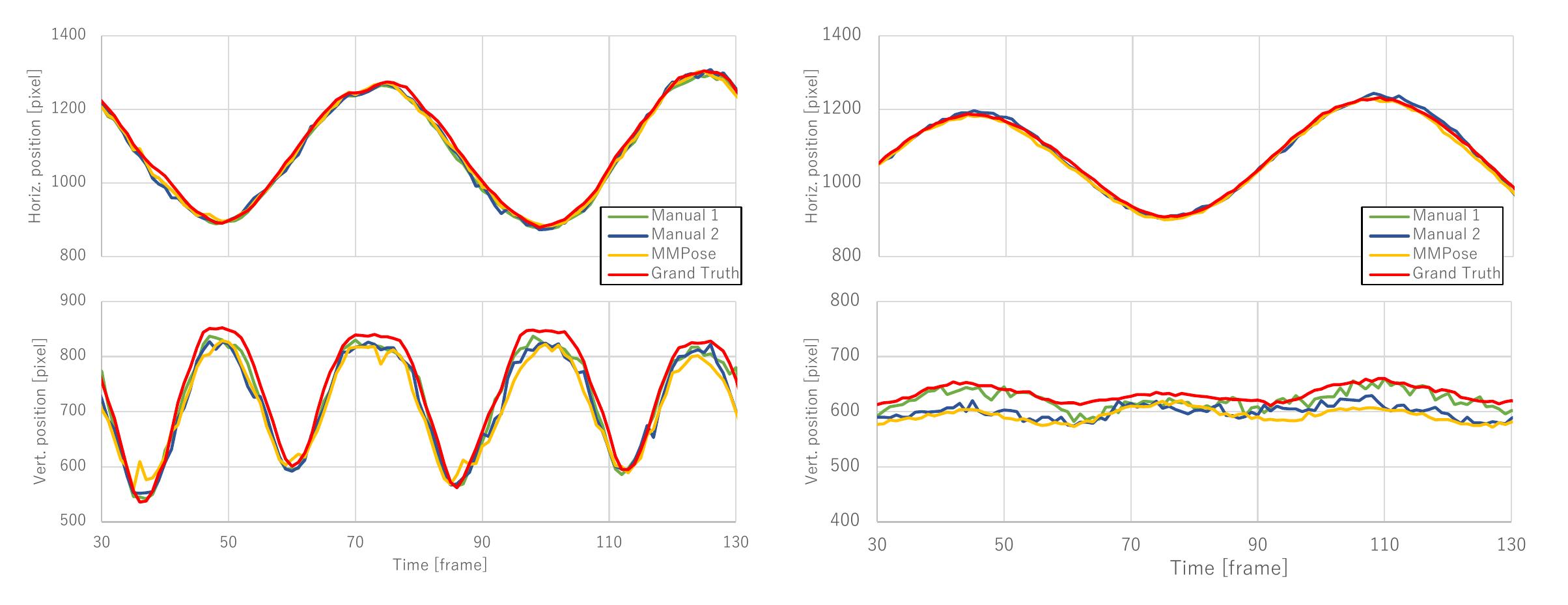}
  \end{center}
  \caption{The resultant position of the left knee for subject A; the top images show the horizontal position, and the bottom images show the vertical position. The left images show the result for the sideways jump, and the right images show that of the ski simulator.}
  \label{fig:fig6}

  \begin{center}
    \includegraphics[pagebox=cropbox,width=0.48\textwidth]{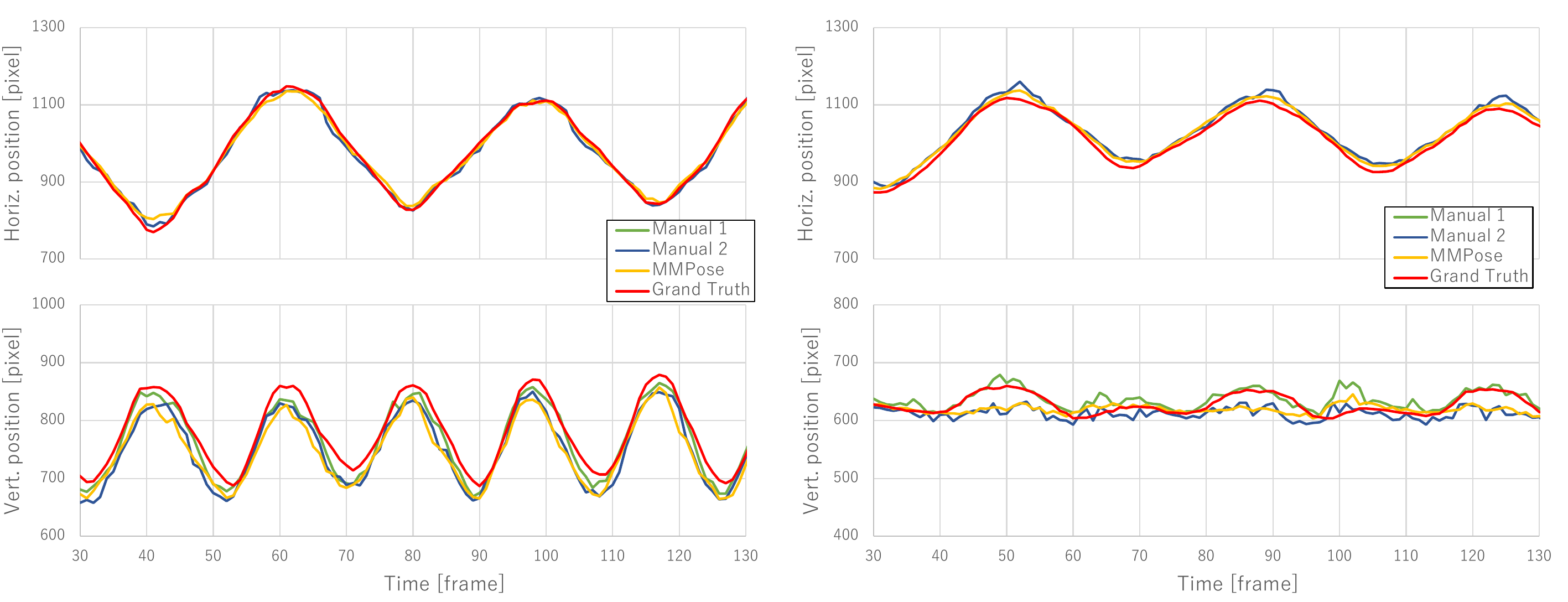}
  \end{center}
  \caption{The resultant position of the left knee for subject B.}
  \label{fig:fig7}
\end{figure}

Figures \ref{fig:fig6} and \ref{fig:fig7} show the results of the left knee; Fig. \ref{fig:fig6} shows the results for the first subject, while Fig. \ref{fig:fig7} depicts the results for the second subject. The left side of each figure depicts the result of sideways jumping, and the right side depicts the ski motion. As shown in Figs. \ref{fig:fig6} and \ref{fig:fig7}, the graphs exhibit similar trajectories.

\begin{figure}
  \begin{center}
    \includegraphics[pagebox=cropbox,width=0.32\textwidth]{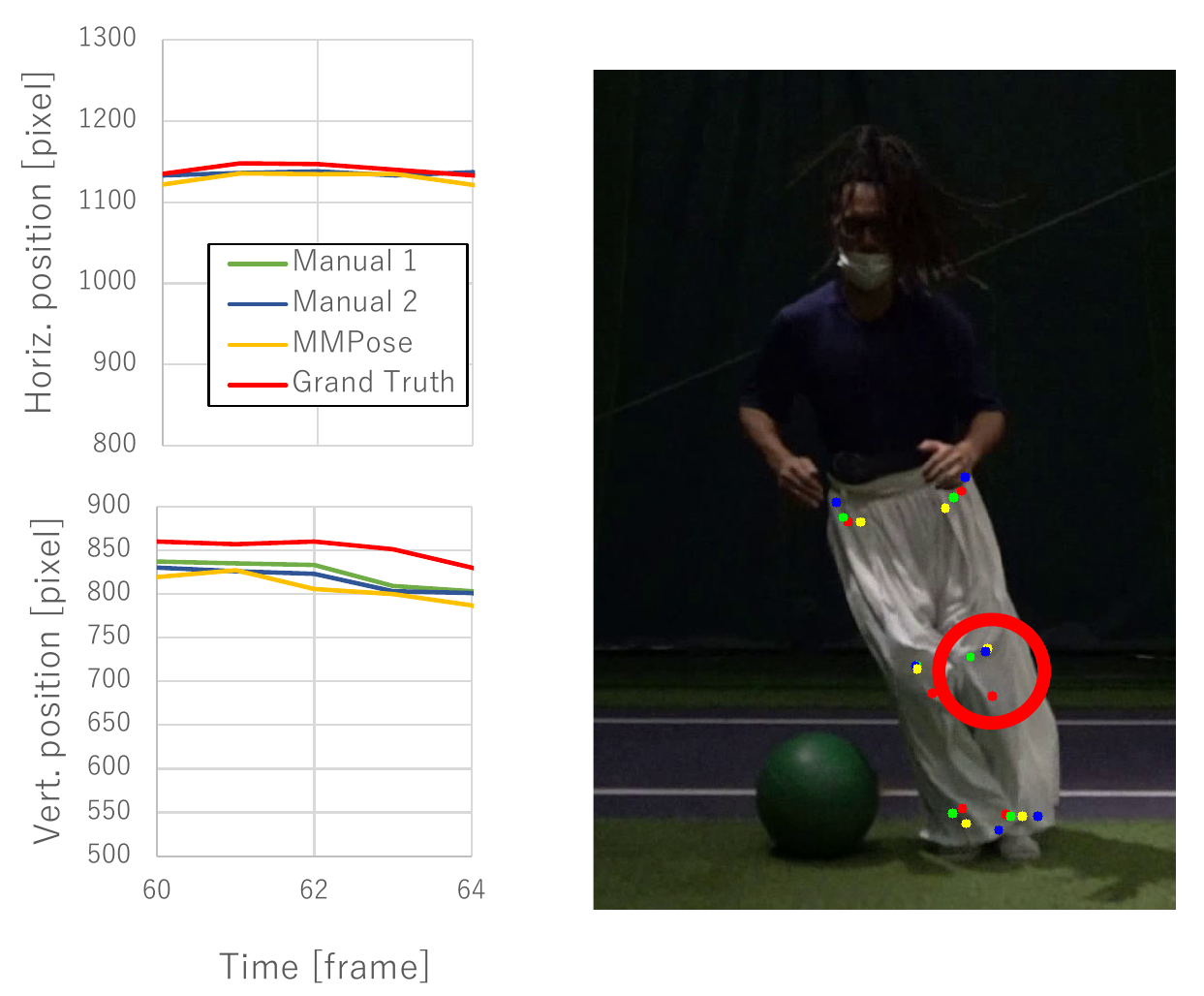}
  \end{center}
  \caption{Example of erroneous result: bulge is considered as knee position.}
  \label{fig:fig8}
\vspace{-0.5cm}
  \begin{center}
    \includegraphics[pagebox=cropbox,width=0.32\textwidth]{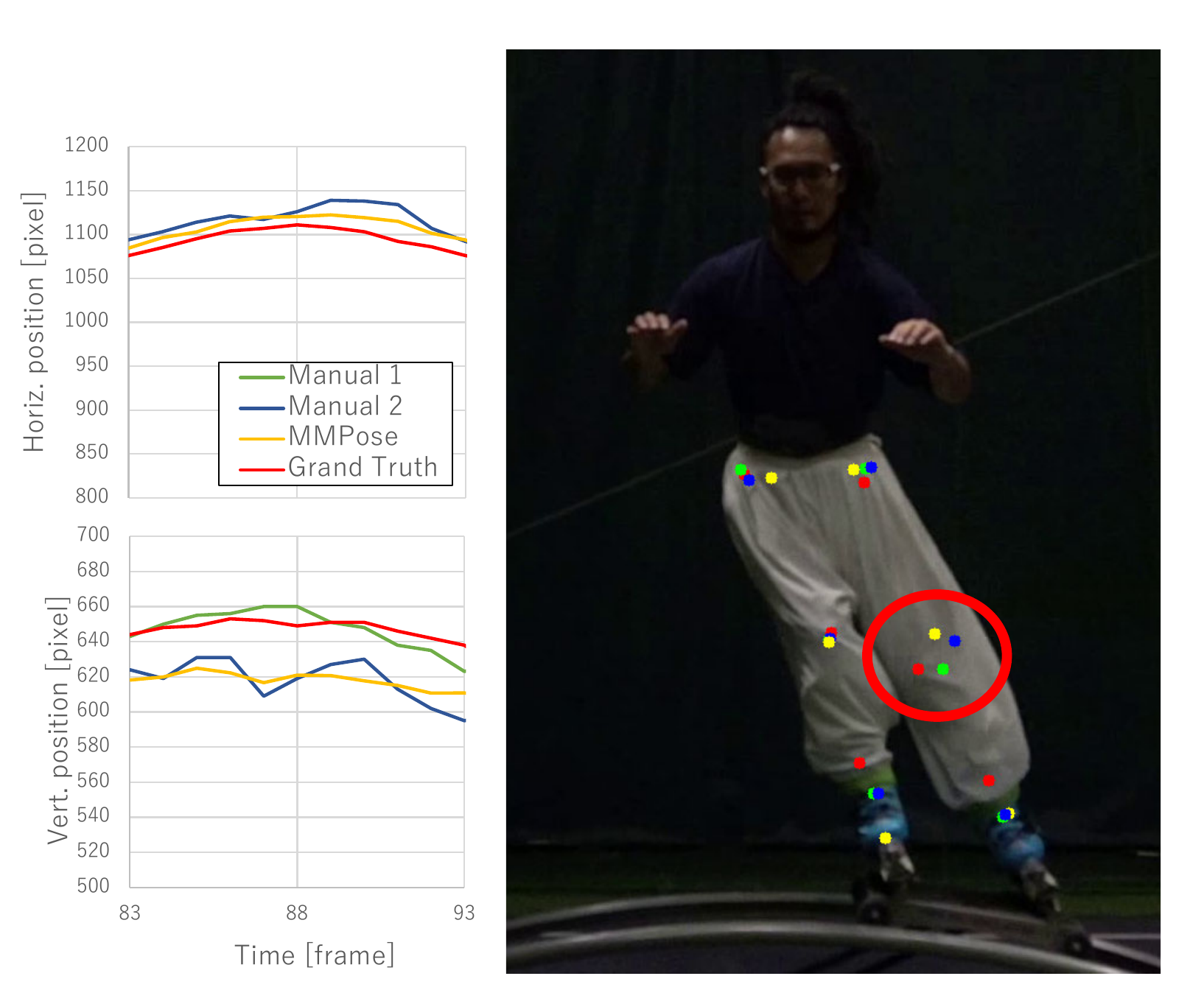}
  \end{center}
  \caption{Example of erroneous result: center of clothes is considered as knee position.}
  \label{fig:fig9}
\end{figure}

However, detailed observations reveal certain differences. Figure \ref{fig:fig8} shows an example of an inaccurate estimation during sideways jumping. In Fig. \ref{fig:fig8}, green and blue dots represent the positions annotated by human operators, whereas orange and red dots denote the positions annotated by MMPose and the ground truth, respectively. Figure \ref{fig:fig8} shows the likelihood of human operators making a similar mistake; the bulge around the knee creates the illusion of indicating the knee position, which is different from the ground truth knee position.


Figure \ref{fig:fig9} shows an example of this aspect during ski motion. As shown in Fig. \ref{fig:fig9}, the human operators and MMPose tend to consider that the knee exists at the center of the clothes. However, in this instance, the subject changed the direction of motion (left-to-right to right-to-left), and the loose-fitting clothes kept moving from left to right by inertial law.

Here, our emphasis was on highlighting the difficulty that the human operators faced while working with loose-fitting clothes. Thus, we confirmed that obtaining training data from human operators for machine learning is difficult.

\section{Conclusion and Future work}

This paper proposes a novel method for obtaining ground truth data for loose-fitting clothes. And we verified the accuracy of existing methods, which strongly supports the necessity of our work. 



Unlike most existing pose estimation frameworks, the proposed method places significance on temporal information. Therefore, in future research, we intend to use a network architecture that can handle time-intensive data. We will prepare various sequences involving loose-fitting clothes as ground truth data and use them to train a deep learning framework.


{\small
\bibliographystyle{ieee_fullname}
\bibliography{egbib}
}

\end{document}